\documentclass[letterpaper, 10 pt, conference]{ieeeconf}  % Comment this line out if you need a4paper

\IEEEoverridecommandlockouts                              % This command is only needed if 
\usepackage{xcolor}

\usepackage{amsmath}
\usepackage{graphicx}
\usepackage{subcaption}
\title{\LARGE \bf
Ensuring UAV Safety: A Vision-only and Real-time Framework for Collision Avoidance Through Object Detection, Tracking, and Distance Estimation
}

\author{Vasileios Karampinis$^{1}$, Anastasios Arsenos$^{2}$, Orfeas Filippopoulos$^{3}$, 
Evangelos Petrongonas$^{4}$, Christos Skliros$^{5}$,\\ Dimitrios Kollias$^{6}$,  Stefanos Kollias$^{7}$, Athanasios Voulodimos$^{8}$
\thanks{$^{1}$Vasileios Karampinis is with the School of Electrical \& Computer Engineering, National Technical University Athens, Polytechnioupoli, Zografou, 15780, Greece (vkarampinis@ails.ece.ntua.gr)}
\thanks{$^{2}$Anastasios Arsenos is with the School of Electrical \& Computer Engineering, National Technical University Athens, Polytechnioupoli, Zografou, 15780, Greece (anarsenos@ails.ece.ntua.gr)}
\thanks{$^{3}$Orfeas Filippopoulos is with the Hellenic Drones S.A., Grigoriou Lampraki 17, Piraeus, 18533, Greece (o.filippopoulos@hellenicdrones.com)}
\thanks{$^{4}$Evangelos Petrongonas is with the School of Electrical \& Computer Engineering, National Technical University Athens, Polytechnioupoli, Zografou, 15780, Greece (vpetrog@microlab.ntua.gr)}
\thanks{$^{5}$Christos Skliros is with the Hellenic Drones S.A., Grigoriou Lampraki 17, Piraeus, 18533, Greece}
\thanks{$^{6}$Dimitrios Kollias is with the School of Electronic Engineering \& Computer Science, Queen Mary University of London, UK (d.kollias@qmul.ac.uk)}
\thanks{$^{7}$Stefanos
Kollias is with the School of Electrical \& Computer Engineering, National Technical University Athens, Polytechnioupoli, Zografou, 15780, Greece (stefanos@cs.ntua.gr)}
\thanks{$^{8}$Athanasios Voulodimos is with the School of Electrical \& Computer Engineering, National Technical University Athens, Polytechnioupoli, Zografou, 15780, Greece (thanosv@mail.ntua.gr)}}

\usepackage{multirow}
\begin{document}

\maketitle
\thispagestyle{empty}
\pagestyle{empty}

%%%%%%%%%%%%%%%%%%%%%%%%%%%%%%%%%%%%%%%%%%%%%%%%%%%%%%%%%%%%%%%%%%%%%%%%%%%%%%%%
\begin{abstract}

In the last twenty years, unmanned aerial vehicles (UAVs) have garnered growing interest due to their expanding applications in both military and civilian domains. Detecting non-cooperative aerial vehicles with efficiency and estimating collisions accurately are pivotal for achieving fully autonomous aircraft and facilitating Advanced Air Mobility (AAM). This paper presents a deep-learning framework that utilizes optical sensors for the detection, tracking, and distance estimation of non-cooperative aerial vehicles. In implementing this comprehensive sensing framework, the availability of depth information is essential for enabling autonomous aerial vehicles to perceive and navigate around obstacles. In this work, we propose a method for estimating the distance information of a detected aerial object in real time using only the input of a monocular camera. In order to train our deep learning components for the object detection, tracking and depth estimation tasks we utilize the Amazon Airborne Object Tracking (AOT) Dataset. In contrast to previous approaches that integrate the depth estimation module into the object detector, our method formulates the problem as image-to-image translation. We employ a separate lightweight encoder-decoder network for efficient and robust depth estimation. In a nutshell, the object detection module identifies and localizes obstacles, conveying this information to both the tracking module for monitoring obstacle movement and the depth estimation module for calculating distances. Our approach is evaluated on the Airborne Object Tracking (AOT) dataset which is the largest (to the best of our knowledge) air-to-air airborne object dataset.

\end{abstract}

%%%%%%%%%%%%%%%%%%%%%%%%%%%%%%%%%%%%%%%%%%%%%%%%%%%%%%%%%%%%%%%%%%%%%%%%%%%%%%%%
\section{INTRODUCTION}

Concerns regarding mid-air collision (MAC) and near mid-air collision (NMAC) are significant in both manned and unmanned aircraft operations, especially in low-altitude airspace. Sense and avoid, refers to an aircraft's capability to maintain a safe distance from and avoid collisions with other airborne traffic. In conditions adhering to visual flight rules, pilots mitigate NMAC/MAC threats by visually detecting and avoiding other aircraft to ensure safe separation. For medium to large airborne systems, active onboard collision avoidance systems such as the Traffic Alert and Collision Avoidance System or the Airborne Collision Avoidance System rely on transponders in cooperative aircraft. However, not all airborne threats can be tracked using transponders, presenting challenges for reliable operations in scenarios involving rogue drones, gliders, light aircraft, and inoperative transponders.

Ensuring aviation safety is paramount. Human vision acts as the final line of defence against mid-air collisions, underscoring its critical role in aviation safety. To aid pilots in mitigating mid-air collision risks, machine vision can be employed to provide alerts regarding potential aircraft and objects in the airspace. Radar usage is often impractical due to the size, weight, and power (SWaP) limitations of Unmanned Aerial Systems (UASs). As a result, machine vision, utilizing CNN-based networks, has emerged as a promising avenue of research to address these challenges \cite{airspacesurvey}.

Machine vision is a widely explored area in onboard systems, enabling machines to perceive their surroundings. With the rapid advancement of computer vision, machine vision has emerged as a promising technology for identifying potential threats [6]. Various approaches exist for object detection, including one-stage \cite{yolox2021, yolov5} and multi-stage detection \cite{detr, chen2023diffusiondet, fasterrcnn} pipelines. Deep learning, in particular, has gained significant traction in machine vision for its capabilities in object detection, tracking \cite{strongsort}, and depth estimation \cite{depthsurvey}.

Cutting-edge approaches \cite{bigdepth, shimada2022pix2pix} commonly employ Convolutional Neural Networks (CNNs) to extract features for predicting depth values per pixel, surpassing classical techniques \cite{classicdepth} by a wide margin. However, these methods rely on intricate and deep network architectures, leading to substantial computational overheads and impractical real-time execution without high-end GPUs. Consequently, deploying such methods on time-sensitive platforms like small drones becomes unfeasible. Conversely, \cite{dong2022mobilexnet} proposed an encoder-decoder network using a lightweight CNN. 

In this work, we present a deep-learning solution that integrates 
vision-only detection, tracking and depth estimation to facilitate real-time conflict detection of non-cooperative aerial vehicles. By non-cooperative aerial vehicles, we denote UAVs for which the sole information is provided by the employed sensor with no other supplementary information supplied from external sources like Radars. More particularly we propose,

\begin{itemize}
    \item Using the distance information (GPS) provided in the AOT dataset we construct a large depth estimation dataset suitable for training an encoder-decoder deep neural network for depth estimation. 
    \item  We design a hybrid loss function to train the aforementioned depth estimation model 
    \item  We integrate our depth estimation model into the detection and tracking pipeline of \cite{arsenos4674579nefeli} and evaluate its performance on a large Airborne Object Tacking dataset achieving remarkable accuracy
     
\end{itemize}

The paper is structured as follows. Section \ref{sec:related} offers a comprehensive literature review of the vision-based methods for the sense and avoidance of UAVs and previous works on depth estimation. In Section \ref{sec:probl}, the problem formulation is outlined, presenting the principal contributions of each element within the proposed pipeline. Section \ref{sec:dataset} provides an in-depth analysis of the dataset configuration that is used to train the depth estimation component.
In Section \ref{sec:expsetup} we outline the experimental setup. Section \ref{sec:evalresults} focuses on the outcomes of the proposed deep learning depth estimation model. Section \ref{sec:concl}, provides the concluding remarks and the way forward to enhance the capabilities of our pipeline.

\section{Related work}
\label{sec:related}

\subsection{Deep learning-based sense and avoid}

In terms of passive sensors, visual cameras are favoured due to their beneficial balance between size, weight, power (SWaP) constraints, and sensing capabilities. Deep learning (DL) techniques have recently made significant strides in addressing intruder detection tasks, as demonstrated in \cite{ieeetranssenseandavoid} and \cite{airtrack}, where comprehensive datasets containing various UAV \cite{AOT, detFly} types and the performance of diverse DL-based detection methods are outlined. 

In \cite{AirtoAirMAV}, a deep learning-based detector integrated with a Kalman filter tracker is proposed to enable real-time air-to-air Micro Air Vehicle (MAV) detection and tracking using an onboard processing
unit and a single camera. In \cite{OPROMOLLA} visual detection involves the utilization of two Deep Learning-based neural networks, each tasked with operations above and below the horizon, respectively. Following this, local image analysis is employed to enhance the precision of the intruder's position detected on the image plane. Subsequently, tentative track generation and firm tracking procedures are conducted, leveraging local association, multi-temporal frame differencing, and Kalman filtering techniques. Lastly, conflict detection is applied to each firm track, utilizing estimates of the line of sight and line of sight rate in stabilized coordinates.
Furthermore, Arsenos et al. \cite{arsenos4674579nefeli} present a visual airborne Sense and Avoid (SAA) system tailored for fixed-wing UAV detection, focusing on deep learning detection and tracking components. Under optimal detection conditions, which necessitate good visibility and unobstructed backgrounds (e.g., clear sky), detection ranges of up to kilometres are achievable (for relatively large targets like a Cessna aircraft).

\subsection{Long distance depth estimation}

In order to effectively address airborne obstacle avoidance, information regarding the obstacle's distance is necessary. However, relying solely on a single camera for distance estimation presents challenges. Some methods utilize known parameters such as the camera's focal length and the object's height to compute distance using the pinhole model, assuming prior knowledge of object dimensions \cite{depthknowniros}. Alternatively, deep-learning approaches have been explored for distance estimation. Xu et al. \cite{aerospace2023distance}, calculate the relative distance by performing reliable 6D pose estimation by solving the Perspective-n-Point (PnP) problem using the feature detection outcomes obtained from the air-to-air image. However, installing a high-precision processor on the UAV is necessary. Furthermore, the current feature encoding method may not be suitable when two UAVs mutually interfere with each other in the same image.

Another approach, presented in \cite{scadinavian, airtrack}, models the depth estimation problem as an image-to-number regression task and integrates it into an object detector. They utilize bounding box information from objects detected by YOLO to estimate distances between objects and the camera without explicit camera parameters or scene knowledge. However, the accuracy of distance estimation may be affected by the variability in bounding box dimensions and generally the problem formulation as a regression task is prone to bias.

As indicated in \cite{depthsurvey}, a real-time monocular depth estimation with accuracy and efficiency balance can be achieved using a lightweight encoder-decoder architecture. In this setting, the deep learning model is trained to translate the input image from the monocular camera to a depth mask, classifying every pixel into depth values.

\section{Problem Formulation}
\label{sec:probl}

\begin{figure*}
    \centering    \includegraphics[width=1.0\textwidth]{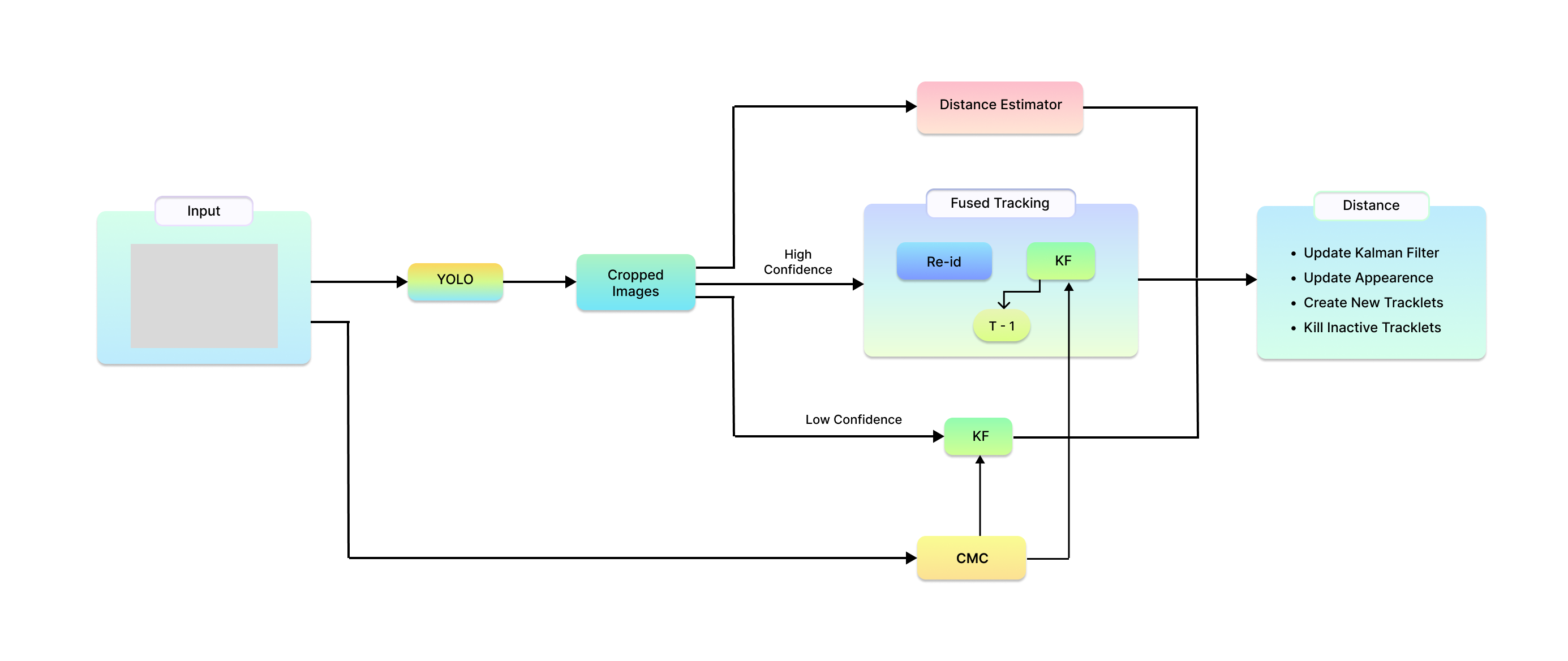}
    \caption{Our proposed system's pipeline}
    \label{fig:pipeline}
\end{figure*}

Our system which is an extension of the NEFELI pipeline \cite{arsenos4674579nefeli} consists of multiple models (demonstrated in Fig. \ref{fig:pipeline}), each specifically designed to effectively address the unique characteristics of detecting,  tracking and estimating the distance of airborne objects.

The proposed workflow initiates with the input image undergoing processing through the detection module. Following this, image sections containing aircraft bounding boxes are cropped to ensure that the bounding box remains centered within the cropped image, known as BBOX Cropping. Post-cropping, the images containing the detected aircraft are processed in the tracking component. During this phase, high-confidence detections are managed by the appearance Re-ID model, whereas low-confidence detections are handled by the Kalman filter-based (KF) motion model. Importantly, the KF motion model integrates camera motion compensation (CMC) effects to account for the influence of camera movement. The tracking module updates the state of either the appearance model or the motion model, generating new tracklets to sustain the tracking of an aircraft or eliminate inactive tracklets. 

In parallel with the tracking procedure, the images containing the detected aircraft are processed in the depth estimation model. More particularly, these images are passed through the encoder network, which extracts hierarchical features by progressively reducing the spatial dimensions while increasing the number of channels. These encoded features capture various levels of abstraction, including edges, textures, and object shapes. Subsequently, the encoded features are fed into a decoder network, which upsamples the feature maps to the original resolution of the input image. The decoder produces a depth map as the output, where each pixel corresponds to the estimated distance of the corresponding pixel in the input image from the camera. The final outcome of the whole pipeline comprises the tracked aircraft, inclusive of the bounding box of the detected aircraft, its associated track ID, and depth estimation information.

\section{Dataset Configuration}
\label{sec:dataset}

We conducted our training procedure with the help of Airborne Object Tracking (AOT) \cite{AOT} dataset. AOT dataset is a collection of approximately 5,000 flight sequences captured from aerial platforms such as drones, helicopters and other air vehicles. These sequences depict diverse environments of urban and natural landscapes. The provided sequences are accompanied by a comprehensive list of annotations including the bounding boxes of the tracked objects. The dataset also provides some contextual information about the video sequences. Such information includes the distance of the object of interest, the geographic coordinates and camera parameters.   

For our study, we dismantled these sequences into frames and organized a dataset based on these images. We also employed the bounding boxes annotations and distance metadata to create our ground truths for the depth estimation problem. Depth estimation datasets \cite{Silberman:ECCV12} generally comprise of image pairs, one corresponds to the image used for training and the other to the depth map whose values equate to the distance of the objects from the camera.        

AOT dataset does not contain such annotations emphasizing the necessity to leverage provided information to construct depth maps (depth annotations). For this purpose, we employed the bounding box information and the distance of the object of interest from the camera. AOT determines the object distance through the employment of GPS technology, given that the coordinates of the object we want to predict and the object containing the camera are known, the distance is ascertained as the difference between these two coordinates meaning that the information the AOT dataset provides refers to the distance of the objects in a straight line.

Utilizing these provided information we created a depth map where the values contained by the bounding box of the object of interest are all equal to the provided distance of the object from the camera. On the contrary, for the areas outside the bounding box, we assign values equal to the maximum distances measured in the AOT dataset.

This approach provides us with the needed ground truths to train our encoder-decoder depth estimation model. We are forced to make assumptions that may not completely adhere to the real data. One such assumption is that the bounding box strictly encloses information related to the object of interest, thereby excluding areas depicting the background. Nevertheless, the evaluation results of our approach on the AOT dataset showcase that the bias produced from these assumptions is negligible.

\section{Experimental Setup}
\label{sec:expsetup}

\subsection{Benchmark (AOT)}

In most cases, depth estimation \cite{MING202114} is defined as an image-to-image regression problem and is typically handled by creating frameworks that can accurately predict a map with the same size as the input image and values representing the distance of each object. The training occurs by minimizing the distance between the predicted map and the ground truth target.

\begin{figure}[b]
    \centering
    \includegraphics[width=0.4\textwidth]{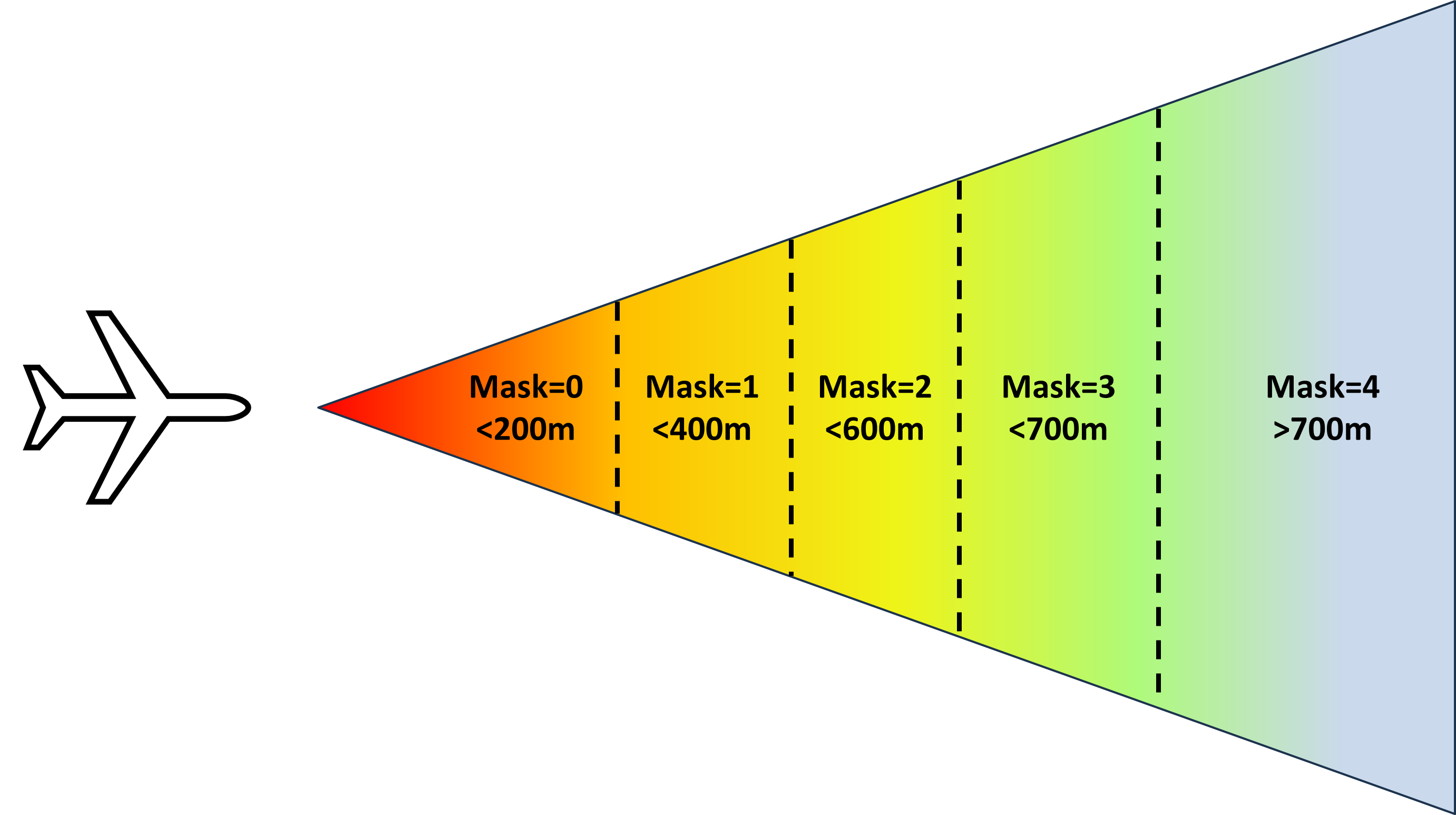}
    \caption{Collision avoidance/safe separation thresholds}
    \label{fig:threshold}
\end{figure}

Other works \cite{cao2017estimating} suggest that a classification approach to the depth estimation problem can be more beneficial. Especially, in the UAV sense and avoid scenario where the distances are very long, it is suggested to use collision avoidance/safe separation thresholds \cite{categoriestut} (see Fig. \ref{fig:threshold}). For this scenario,  the problem is formulated as a multiclass classification task with N depth group increments($d_{0}, …, d{1}$). Each grouped increment corresponds to a distinct class within the classification scheme. 

In this work, we follow the classification framework. Based on the distance information we categorize the data into 4 different classes and we define one last class as the background. 
\begin{itemize}
    \item First class: objects with a distance value less than 200 meters
    \item Second class: objects with a distance value less than 400 meters
    \item Third class: objects with a distance value less than 600 meters
    \item Fourth class: objects with a distance value less than 700 meters
    \item Fifth class: background objects with a distance value over 700 meters
\end{itemize}

Following the formula described above we created our new ground truths \textit{Masks} for the training procedure which are $W\text{x}H$ arrays with values varying from 0-4

\begin{equation}
        \textit{Mask} =
    \begin{cases}
        0, & \text{if distance} < 200 \text{ meters} \\
        1,  & \text{if distance} < 400 \text{ meters}\\
        2, & \text{if distance} < 600 \text{ meters} \\
        3,  & \text{if distance} < 700 \text{ meters}\\
        4,  & \text{if } \text{distance} > 700 \text{ meters} 
    \end{cases}
    \quad N\text{x}N
\end{equation}  

We intend to train an image-to-image model with an output close to the mask labels. Although typical classification problems achieve this by minimizing cross-entropy loss, applying such a loss to our problem is inefficient due to its nature. Initially, our data are ordered, and for this reason, wrong predictions in samples categorized into classes representing longer distances should be punished more compared to the closer ones. Additionally, our labels are occupied by 4 values for the most part given the small size the point of interest accounts for the whole image. This saturation of 4 values creates a bias in the model, leading to undesirable results.

In order to address the added bias problem, the input images were prepossessed by cropping around the center of the area of interest, which refers to the provided bounding box. With this center crop around the point of interest, we manage to constrain the bias and remove irrelevant information for training the distance estimation model. Additionally, a Gaussian filter was applied to the center-cropped created masks to mitigate the significant value divergencies at the edges of the area of interest and provided a normalization which is beneficial during training. Lastly, we exploited and combined the benefits of different losses like structural similarity index \cite{ssim}, edge-based losses, L1 and Berhu which was popularized in \cite{laina2016deeper} and is commonly used in studies tackling the depth estimation task \cite{l2seeda} to train a strong-performing, robust model that is capable of providing satisfactory results in domain shifts \cite{ben-david2010theory}.
\begin{itemize}
    \item Edge loss retains the boundaries of an object by penalizing false predictions in the edge of the object more harshly than in other areas.
    
    \item Structural Similarity index measures the degree of similarity between two images by utilizing luminance similarity, contrast similarity and structural similarity. These similarities are commonly calculated with the utilization of sliding windows. 

    \item Berhu loss is a robust regression loss function used in depth estimation tasks and It is designed to mitigate the influence of outliers in the training data while maintaining sensitivity to small errors.
\end{itemize} 

\begin{figure*}[htb]
    \centering
    \includegraphics[width=0.7\linewidth]{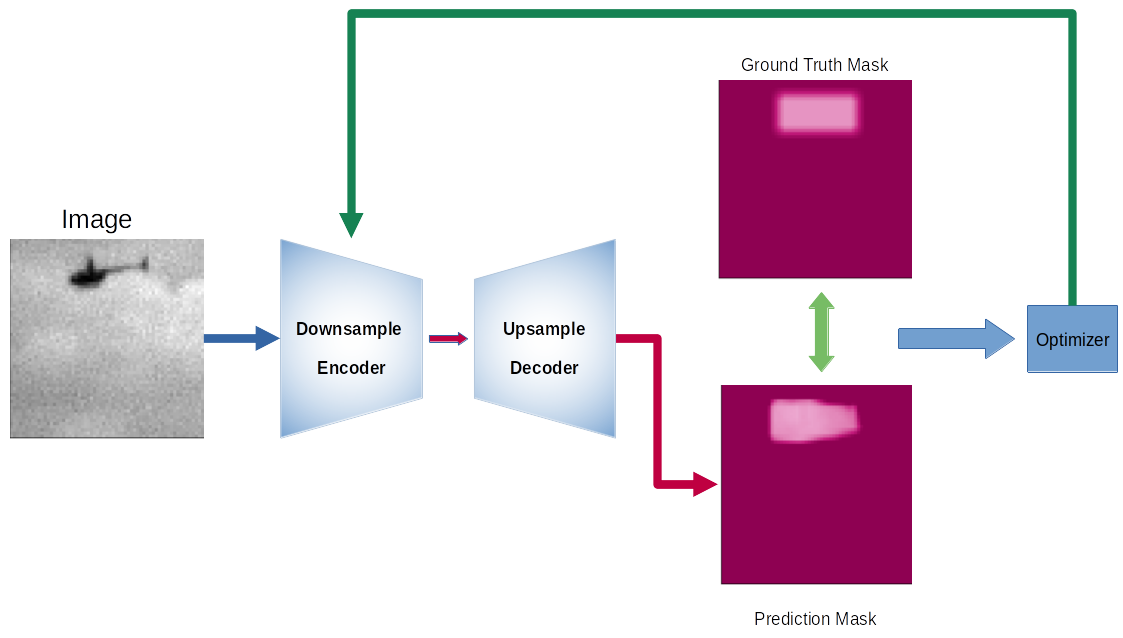}
    \caption{Pipeline of the proposed depth estimation model}
    \label{fig:Depth estimation pipeline}
\end{figure*}

For the training procedure, we implemented a U-net convolutional neural network \cite{ronneberger2015unet}. The U-net architecture comprises an encoder and a decoder block. The encoder's convolutional blocks reduce spatial dimensions and increase the depth of feature maps to encapsulate higher-level features. The decoder block upsamples the extracted feature maps from the encoder block to restore the lost spatial resolution. For our training framework, we chose Adam \cite{kingma2017adam} as our optimizer with a weight decay equal to 0.0005 and an adaptive learning rate given by Eq.\ref{eq:warmup learning rate} to enhance the stability of the training convergence.  To assist in the improvement of the model in generalization and further mitigate the inserted bias we utilized L2 regularization \cite{l2reg} which encourages the model to distribute weight more evenly across all features by penalizing the squared magnitudes of the coefficients in the loss function. This technique incentivizes the model to prioritize coefficients that contribute proportionately across all features, promoting smoother decision boundaries and reducing the model's susceptibility to overfitting. The overall architecture of the depth estimation model is showcased in Fig. \ref{fig:Depth estimation pipeline}, where an image is provided as an input to the encoder-decoder model then the model produces a prediction mask. The purpose of the model is to generate a prediction mask that mirrors its corresponding ground truth counterpart.

\begin{equation}
    \label{eq:warmup learning rate}
        f(x) = 
    \begin{cases}
        1, &\text{\small if } x \geq \text{\small warmup iterations}, \\ 
        \gamma * (1-\alpha) + \alpha, &\text{\small if } x < \text{\small warmup iterations}
    \end{cases}
\end{equation}
\begin{align*}
\text{where } \alpha &= \frac{x}{\text{warmup iters}}, \gamma = 0.001, \\
\text{warmup iterations} &= \text{min}(1000, \text{length(dataset)}-1)
\end{align*}

% \subsection{Real flights (Generalisation Ability)}

% In the evergrowing field of aviation where a clear view of the airborne traffic should be maintained the need to develop depth estimation models that can perform well on unseen data is crucial. For this reason, we tested our previously trained models in real-flight videos and discerned the model's performance in these new domain-shifted data. In the evaluation section of this study, we demonstrate and analyze the results of the framework in some keyframes of a benchmark video. We conducted this experiment in an approximately one-minute video with dimensions 1200x800, where we first predicted the displayed airborne objects in the image and then estimated the distance of the detected object based on the classification bins showcased in the Benchmark(AOT) section. 

\section{Evaluation results}
\label{sec:evalresults}

A widely utilized and pragmatic metric for evaluating the efficacy of depth estimation is either the mean absolute error or the root mean square error. In consideration of the manner in which we have delineated our problem regular regression metrics might not be quite enlightening in discerning the performance of our model, for this reason, we proposed four metrics that could be used in the classification approach of the depth estimation task. These metrics provide a classification value that conveys precision in the categorization of each detected object in the correct classification bin. More specifically we utilized sliding a window across the area of interest (bounding box) to extract spatial information about the predicted distance of the object and employed methods like mean, max and min to combine the retrieved windows. Another, metric we applied was the rate of similarity between two pictures where the classification is considered truthful when the rate surpasses a given threshold. The metrics mentioned are expounded upon in the subsequent section below.

\subsection{Evaluation Results on the AOT Dataset}
In this section, we present and analyze the results of the conducted experiments for the different tested components. We also, analyze how each component affects the different losses on the overall performance of the introduced metrics. First, we trained our model applying only the edge loss function. Let $Y(x,y)$ be the predicted map and $I(x,y)$ be the input image, we calculate the edge loss by retrieving the gradients of the predicted map array in both the x and y-axis. After assessing these gradients we apply a smoothness factor with the help of the input image $I(x,y)$ and obtain the loss value as the sum of the mean absolute values for the x and y axis respectively as shown in Eq.\ref{eq:edge loss}

\begin{equation}
    \label{eq:edge loss}
    EL_{x,y} = e^{- \frac{1}{N}\sum_{i=0}^{i=N} \left| \nabla_{\hat{x},\hat{y}}I(\hat{x},\hat{y}) \right|}\cdot\nabla_{x,y}Y(x,y)
\end{equation}
  
\begin{figure}[htb]
    \centering
    \includegraphics[width=1.0\linewidth]{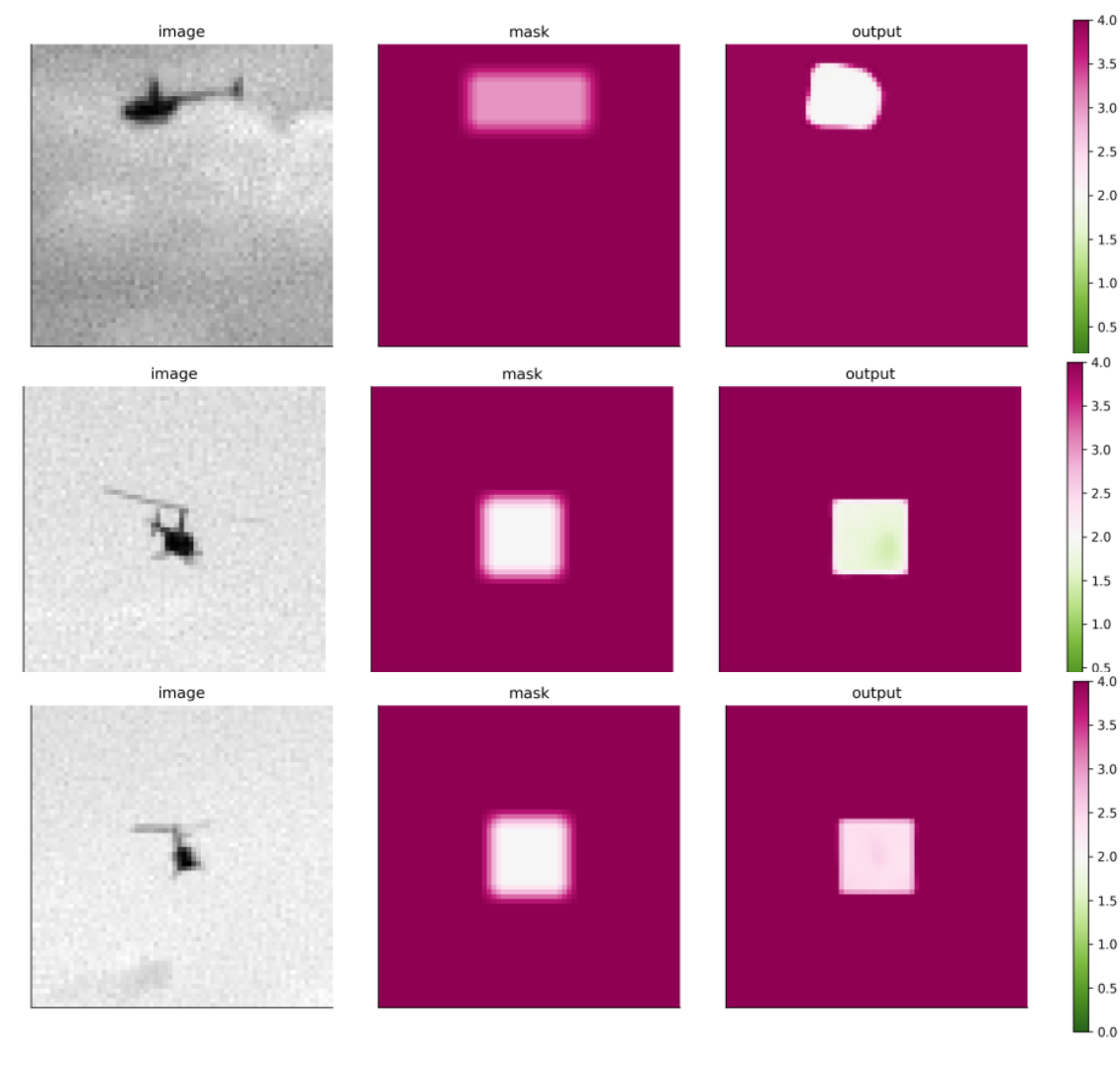}
    \caption{Depth estimation visualization for L1, Berhu and multi loss respectively}
    \label{fig: Losses estimation}
\end{figure}

The model trained with this loss achieves poor results depicted in Table \ref{tab:reg_loss_table}. It's reasonable to anticipate this outcome, considering its intended function to safeguard the structural intricacies and boundaries found within the input images and depth maps. Nevertheless, edge loss remains a pivotal element within the domain of depth estimation, as it contributes to the generation of depth maps characterized by enhanced smoothness and visual coherence, thereby portraying objects with greater clarity and contrast. Next, the model was trained using the L1 loss, as a standard loss function in regression tasks. From the regression metrics we can observe that models trained with this loss function are capable of producing satisfactory results. From Fig. \ref{fig: Losses estimation}, it could be deciphered that the L1 function trains the model to sufficiently pinpoint the area of interest but the outliers of the area are not adequately defined.

\begin{figure*}[htb]
    \centering
    \includegraphics[width=1.0\linewidth]{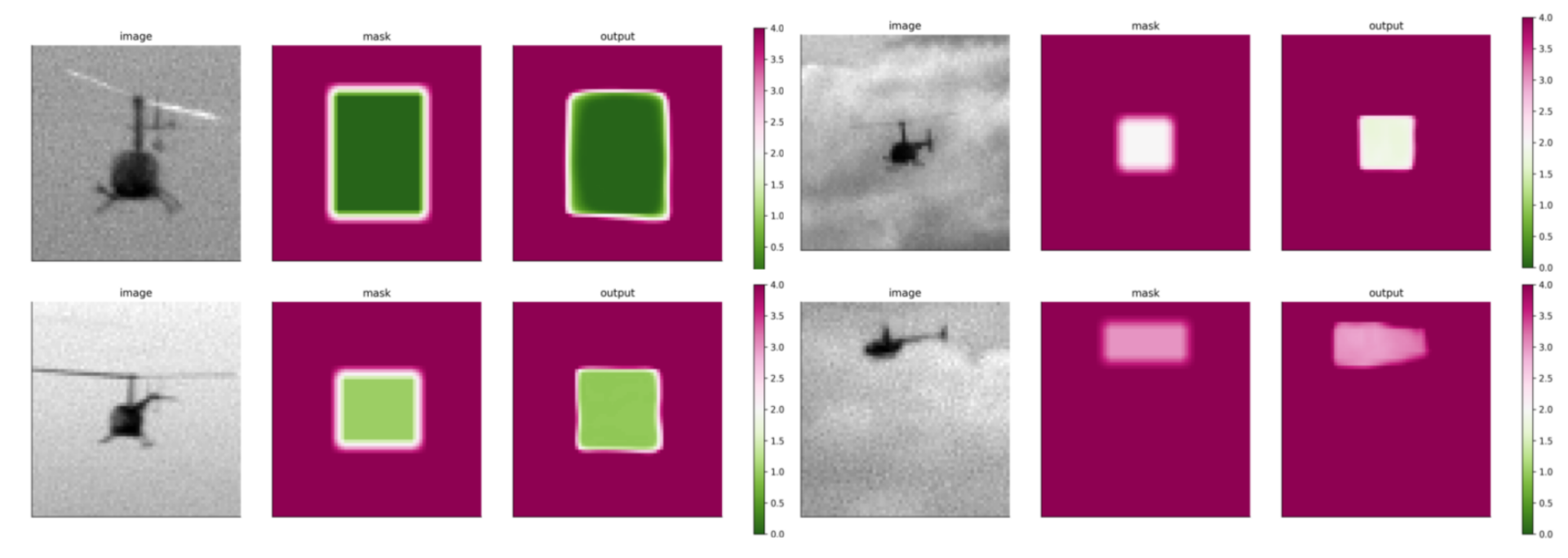}
    \caption{Depth estimation ground truth and prediction mask for each classification bin}
    \label{fig: prediction for each class}
\end{figure*}

For the forthcoming experiment, we assessed the efficacy of the model trained to utilize the BerHu function \cite{laina2016deeper}. This function incorporates the benefits of both the L1 loss and L2 so it is natural to expect a better performance which is something that can be observed in Table \ref{tab:reg_loss_table} as the BerHu loss \cite{laina2016deeper} achieves better performance in both the Mean Absolute Error (MAE or L1)\cite{bispat} and Root Mean Square Error (RMSE) \cite{bispat} loss. 
To assess the viability of the BerHu loss as a substitute for the L1 loss function, we conducted an experiment wherein the model was trained using a combination of both loss functions. Subsequently, we evaluated the model's performance. Our observations indicate that the model's performance remained largely consistent, albeit marginally inferior compared to the model trained exclusively with the BerHu loss function. Lastly, all the above loss functions were combined using a weighted function described in Eq.\ref{eq: weight equation}. Predictions for each of the four classification bins of interest using the multi-loss trained model are presented in Fig. \ref{fig: prediction for each class}

\begin{equation}
\begin{aligned}
    Loss &= W_{\text{edge loss}} \cdot EL(\hat{y},y) \\
         &\quad + W_{\text{ssim}} \cdot SSIM(\hat{y},y) \\
         &\quad + W_{L1} \cdot L1(\hat{y},y) \\
         &\quad + W_{\text{berhu loss}} \cdot Berhu(\hat{y},y)
\end{aligned}
\label{eq:weight-equation}
\end{equation}

% \begin{equation}
%     \begin{split}
%     Loss &= W_{edge\,loss} \cdot EL(\hat{y},y) 
%          &+ W_{ssim} \cdot SSIM(\hat{y},y)
%          &\quad+ W_{L1} \cdot L1(\hat{y},y) 
%          &+ W_{berhu\,loss} \cdot Berhu(\hat{y},y)
%     \end{split}
%     \label{eq: weight equation}
% \end{equation}

\begin{table}[b]
\centering
\caption{Regression metrics for the different losses}
\begin{tabular}{|c|c|cc|}
\hline
\multirow{2}{*}{Model} & \multirow{2}{*}{Loss functions} & \multicolumn{2}{c|}{Regression Metrics} \\ \cline{3-4} 
                       &                                 & \multicolumn{1}{c|}{MAE}      & RMSE    \\ \hline
\multirow{5}{*}{Unet}  & Edge                            & \multicolumn{1}{c|}{3.39}     & 3.55    \\ \cline{2-4} 
                       & L1                              & \multicolumn{1}{c|}{0.19}     & 0.43    \\ \cline{2-4} 
                       & Berhu                           & \multicolumn{1}{c|}{0.12}     & 0.36    \\ \cline{2-4} 
                       & L1/Berhu                        & \multicolumn{1}{c|}{0.13}     & 0.37    \\ \cline{2-4} 
                       & Edge/SSIM/L1/Berhu              & \multicolumn{1}{c|}{0.14}     & 0.39    \\ \hline
\end{tabular}
\label{tab:reg_loss_table}
\end{table}

Except for the regression metrics used in evaluating the trained models, our assessment of accuracy relied on our proposed classification metrics. We initiated our approach by devising a metric centered on a sliding window of dimensions 5x5. This window systematically traversed the designated area of interest (bounding box) within the predicted mask to gather spatial depth information at the specified distance. Subsequently, after traversing the window across the point of interest, we collected a series of \textit{m} kernels, each with dimensions \textit{kxk}, to extract a singular value. For each kernel, we computed the mean of the \textit{$k^{2}$} values, resulting in \textit{m} values. Each value represented the estimated depth of its respective window. Following this step, we explored three distinct approaches to deriving the final prediction from the set of \textit{m} values: utilizing the mean, minimum and maximum values. The resultant values ranged from 0 to 4, subsequently rounded to the nearest integer value (0, 1, 2, 3, or 4) to yield the final prediction.

Let $M$ denote the predicted mask, and $D(x,y)$ denote the depth information at coordinates $(x,y)$. For each window position $(x,y)$ within the bounding box where $(x,y)$ stand for the top left coordinates of the window, the depth $D(x,y)$ value for each kernel is calculated by Eq.\ref{eq: Calculate depth values for each kernel}:

\begin{equation}
    \label{eq: Calculate depth values for each kernel}
    D_{kernel}(x,y) = \frac{1}{k^2} \sum_{i=1}^{k} \sum_{j=1}^{k} M(x+i,y+j) 
\end{equation}

For the sliding windows, we utilized a stride value of 1 and applied padding to the predicted mask by reflecting its values. Let $f$ be the function we applied to get the final prediction of depth (mean, min, max). The final prediction can be described by the Eq.\ref{eq: class prediction}
\begin{equation}
    \label{eq: class prediction}
    \text{class} = \text{round}(f(\sum_{i=1}^{m} D_{i}(x,y)))    
\end{equation}

From Table \ref{tab:cls_loss_table_aot} we can see that the worst-performing model is the one trained only with edge loss. This is expected as we observed the same in the regression metrics table (Table \ref{tab:reg_loss_table}). Of all the three different functions tested in the Sliding Window metric, the worst-performing one is the max function. The decrease in performance can be attributed to the significance of outliers within the function.

In particular, a thorough examination of the L1 prediction mask visualization depicted in Fig. \ref{fig: Losses estimation} demonstrates that the bounding box boundaries predicted by the model using L1 loss function are inadequately defined. Since the background has larger values the max function chooses these as the correct categories diminishing the classification results. The accuracy achieved through the sliding window method utilizing the mean value demonstrates a notably superior performance in comparison to that attained through the max function. This outcome aligns with expectations, as the mean function effectively mitigates the influence of larger outlier values. Notably, among the three functions considered, the min function emerges as the most effective performer. Disregarding the larger values introduced by outliers, the minimum function results in the highest performance among the evaluated methods.

\begin{table}[b]
\centering
\caption{Classification metrics for the different losses}
\begin{tabular}{|c|c|cccc|}
\hline
\multirow{3}{*}{Model} & \multirow{3}{*}{Loss functions} & \multicolumn{4}{c|}{Accuracy Metrics}                                                                                                                              \\ \cline{3-6} 
                       &                                 & \multicolumn{3}{c|}{Sliding Window}                                               & \multirow{2}{*}{\begin{tabular}[c]{@{}c@{}}Threshold \\ Accuracy\end{tabular}} \\ \cline{3-5}
                       &                                 & \multicolumn{1}{c|}{Mean} & \multicolumn{1}{c|}{Max}  & \multicolumn{1}{c|}{Min}  &                                                                                \\ \hline
\multirow{5}{*}{Unet}  & Edge                            & \multicolumn{1}{c|}{0.14} & \multicolumn{1}{c|}{0.14} & \multicolumn{1}{c|}{0.14} & 0.01                                                                           \\ \cline{2-6} 
                       & L1                              & \multicolumn{1}{c|}{0.61} & \multicolumn{1}{c|}{0.16} & \multicolumn{1}{c|}{0.71} & 0.86                                                                           \\ \cline{2-6} 
                       & Berhu                           & \multicolumn{1}{c|}{0.64} & \multicolumn{1}{c|}{0.17} & \multicolumn{1}{c|}{0.76} & 0.86                                                                           \\ \cline{2-6} 
                       & L1/Berhu                        & \multicolumn{1}{c|}{0.66} & \multicolumn{1}{c|}{0.18} & \multicolumn{1}{c|}{0.74} & 0.89                                                                           \\ \cline{2-6} 
                       & Edge/SSIM/L1/Berhu              & \multicolumn{1}{c|}{0.63} & \multicolumn{1}{c|}{0.23} & \multicolumn{1}{c|}{0.72} & 0.86                                                                           \\ \hline
\end{tabular}
\label{tab:cls_loss_table_aot}
\end{table}

We additionally employed a metric, known as threshold accuracy, to assess the classification task. This metric computes the similarity of the pixels between the predicted and ground truth images. For each corresponding pixel pair, if the ratio between the predicted and ground truth values remains under a specific threshold, the classification is deemed accurate. The overall accuracy score is then computed as the percentage of the correctly classified pixels relative to the total number of pixels within the mask. For the showcased results in Table \ref{tab:cls_loss_table_aot} we applied a threshold equal to 1.25. Based on the findings presented in Eq.\ref{eq: class prediction}, it is evident that the models exhibiting the poorest performance solely rely on the edge loss function. Conversely, those achieving higher accuracy levels are the models that integrate a combination of the Berhu and L1 loss functions.

\section{Conclusion and Future Work}
\label{sec:concl}

The proposed models trained in the AOT dataset exhibited promising results in regards to accurately discerning the distance values solely from a single image. This is an endorsement and a motivation to expose our models to images belonging to a shifted domain and test their robustness by assessing their ability to perform reliably under conditions they have not encountered before.  

Furthermore exposing our models to real-world flight tests is paramount to assessing the model's adaptability to adverse weather conditions. Real-world flight scenarios elicit unanticipated constraints and complexities that a supervised environment is incapable of addressing. Assessing a model's performance in authentic flight conditions provides a reassuring evaluation that gives an acumen in the model's aptitude to adapt to unforeseen conditions.   

Deploying a model trained in a benchmark dataset to a realistic environment can be a challenging task. The main reason for this phenomenon is the exposure of the model to an extensive range of data discrepancies that might occur due to different weather phenomena such as rain, fog, clouds, or hardware deficiencies like color quantization, iso noise and others. These data divergencies could be conceptualized as domain shifts occurring in the latent space representation of the data. The domain shift \cite{ben-david2010theory} task encapsulates a widely acknowledged challenge for learning algorithms, often leading to depreciation in performance when faced with conditions divergent from those encountered during training. Addressing these challenges will be the main focus of our future work.

\bibliographystyle{IEEEtran}
\bibliography{IEEEfull}

\end{document}